\documentclass[preprint,12pt]{elsarticle}%
\usepackage[labelsep=period]{caption}
\usepackage[hidelinks]{hyperref}
\usepackage{rotating}
\usepackage[utf8]{inputenc}
\usepackage{amsmath}
\usepackage{amsfonts}
\usepackage{amssymb}
\usepackage{graphicx}
\usepackage{booktabs}
\usepackage{xcolor}
\usepackage{multirow}
\usepackage{float}
\usepackage{natbib}

\usepackage{etoolbox}

\DeclareUnicodeCharacter{2212}{-}

\makeatletter
\def\ps@pprintTitle{%
 \let\@oddhead\@empty
 \let\@evenhead\@empty
 \def\@oddfoot{\footnotesize\itshape
       Postprint submitted to \ifx\@journal\@empty Elsevier
       \else\@journal\fi\hfill}
 \let\@evenfoot\@oddfoot}
\makeatother

\journal{Expert Systems with Applications}
\begin{document}
\begin{frontmatter}

%
%
%
%
%
%
%

\title{Improving importance estimation in covariate shift for providing accurate prediction error}
\author[arcelor]{Laura Fdez-Díaz}
\ead{laura.fernandezdiaz@arcelormittal.com}
\author[uniovi]{Sara Gonz\'alez Tomillo}
\ead{gonzaleztsara@uniovi.es}
\author[uniovi]{Elena Monta\~n\'es\corref{cor1}}
\ead{montaneselena@uniovi.es}
\author[uniovi]{Jos\'e Ram\'on Quevedo}
\ead{quevedo@uniovi.es}
\cortext[cor1]{Corresponding author}
\address[arcelor]{Global R\&D, ArcelorMittal (Spain)}
\address[uniovi]{Artificial Intelligence Center. University of Oviedo at Gij\'on, 33204 Asturias,  Spain \texttt{http://www.aic.uniovi.es}}

	\begin{abstract}
		In traditional Machine Learning, the predictions of the algorithms are based on the assumption that the problem data follows the same distribution in both the training and the test datasets. However, in real world problems this condition does not hold and, for instance, the distribution of the covariates changes whereas the conditional distribution of the targets remains unchanged. If this particular situation takes place, we are facing a covariate shift problem where standard error estimation may be no longer accurate. In this context, the importance is a measure commonly used to alleviate the influence of covariate shift on error estimations. The main drawback is that the importance is not easy to compute. The Kullback-Leibler Importance Estimation Procedure (KLIEP) has been shown to be capable of estimating importance in a promising way. Despite the good performance of this procedure, it fails to ignore target information, since it only includes the covariates information for computing the importance. In this direction, this paper explores the potential improvement in the performance of the method if information about the targets is taken into account in the computation of the importance. Then, a redefinition of the importance arises in order to be generalized in this way. Besides the potential improvement in performance, including target information make possible the application to a real application about plankton classification that motivates this research and characterized by its great dimensionality, since considering targets rather than covariates reduces the computation and the noise in the covariates. The impact of taking target information into account is also explored when Logistic Regression (LR), Kernel Mean Matching (KMM), Ensemble Kernel Mean Matching (EKMM) and the naive predecessor of KLIEP called Kernel Density Estimation (KDE) methods estimate the importance.  The results of the experiments lead to conclude that the error estimation is more accurate using target information when either density or probabilities are involved in the importance computation, and, especially in case of the more promising method KLIEP.
	\end{abstract}
\end{frontmatter}

	\section{Introduction}
Traditional supervised learning makes the assumption that the data follow an unknown distribution $P(X,Y)$ that does not change between training and test dataset when the examples are drawn independently and identically distributed. This means that supervised learning assumes that the training set reflects the probability distribution of the problem and expects the test set to follow the same one; that is, the distribution is assumed not to change. However, in certain real-world problems, this assumption is often violated \citep{quionero2009dataset,storkey2009training}. For instance, this situation, called \textit{dataset shift} \citep{kull2014patterns,moreno2012unifying}, can be seen in cases where the population changes overtime or a process where, due to cost concerns, one of the classes is sampled at a lower rate than it actually appears. In these problems, $P(X,Y)$ changes from the training to test data. These dataset shifts can be characterized and categorized according to how the distribution changes \citep{webb2005application}. A particular case takes place when the distribution of the independent variables, also known as covariates, changes between the training and the test set but the distribution of the target variables remains constant. This is known as \textit{covariate shift}. A real world problem of this kind is plankton classification, which, in one sense, was the origin of this research. Plankton plays a vital role in marine ecosystems and performing good predictions is a clue task. The main pitfall is that the characteristics and life cycles of the different classes of plankton change throughout the year. Thus, the distribution of the classes changes depending on the time of the year the examples are drawn. Obviously, this means that the distribution of the covariates changes between training and test sets, but the target distribution remains unchanged.

The difficulty of coping with this kind of problems is that the performance of machine learning algorithms is highly conditioned by the particular data characteristics of the problem. According to the available data, whatever algorithm of this kind is chosen will yield better or worse results. One of the classic metrics used to measure the performance of the different methods is the distance between the actual target values and the predictions yield by the method. However, one of the main pitfalls found in the dataset shift framework is precisely the error estimation, since conventional ways of error estimation do not work properly because they only guarantee good error estimating under the condition that training and test datasets follow the same distribution \citep{sugiyama2008direct}.

Fortunately, depending on the type of distribution change, there exist different techniques that can be applied to soothe the effects of the shift on the error estimation. In the particular case of covariate shift, the concept of importance arises to be included in the conventional error estimation procedures, leading to what is known as log-likelihood estimators. The fact is that the way of computing this measure is not obvious in data shift, since there is no information about test distribution. Recently, a procedure for estimating the importance \citep{sasaki2015direct} has offered good performance in estimating the importance on data shift, as an alternative to other naive approaches. However, this procedure, as its predecessors, also ignores information about the target, since it only considers the covariates for computing the importance. Moreover, these methods have trouble dealing with real-world problems like the plankton classification problem, which gave origin to this research and whose data have high dimensionality, and covariates may include noise or redundancy, which may dismiss the performance of the error estimation. Our hypothesis is that target information may be a promising source that deserves to be explored for improving the performance of the error estimation though importance. The goal of this paper is precisely to analyse the impact of using this information in existing importance estimation methods, and to prove if its use could also allow for error estimation methods to be applied in complex real world problems such as plankton.

The rest of the paper is organized as follows. Section 2 characterizes the situations when different distribution changes take place. This section also details covariate shift and states the error estimation problem though the importance measure. Section 3 gives an overview of existing methods for estimating the importance. Section 4 elaborates the proposal of including target information in the existing importance estimators. Section 5 explains how to introduce covariate shift into a dataset using the concept of prevalence, due to the lack of datasets available in the literature with this distribution change. Finally, Section 6 documents and analyses the experimental results and opens new lines for future research.

\section{Covariate shift statement}
Covariate shift falls into a paradigm called data shift characterized by the particularity that training and test dataset are drawn by different distributions. There exist different categorizations and discussions about problems with dataset shift \citep{kull2014patterns, moreno2012unifying}. Supervised learning problems are defined by a set of independent variables, $x$, also known as covariates, belonging to a feature space $\mathcal{X}$, which follow a probability distribution $P(x)$, and a target variable, $y$, belonging to an image space $\mathcal{Y}$, which follows a probability distribution $P(y)$. In fact, in this type of domains $\mathcal{X}$ and $\mathcal{Y}$, there also exists a joint probability distribution of the covariates and the target variable, $P(x, y)$, which defines the problem and from where the examples are drawn at random, and for which a function $f:\mathcal{X}\rightarrow\mathcal{Y}$ is learned.

	It is important to understand how the data is generated according to the causal relationship between covariates and the target variable. This causal relationship determines the different kinds of distribution changes that can occur. According to the taxonomy proposed in \citep{fawcett2005response}, we can identify two types of domains: $\mathcal{X} \longrightarrow \mathcal{Y}$ domains, in which the target variable is causally determined by the covariates $x$, and $\mathcal{Y} \longrightarrow \mathcal{X}$ domains, in which the covariates $x$ are directly dependent on the target $y$.

	In $\mathcal{X} \longrightarrow \mathcal{Y}$ domains, the joint probability $P(x, y)$ can be written as $P(y|x)\cdot P(x)$, whereas, in $\mathcal{Y} \longrightarrow \mathcal{X}$ domains, the joint probability can be written as $P(x|y)\cdot P(y)$. A dataset shift is considered to occur when the joint probability changes from the training set to the test set, that is, when $P_{tr}(x,y) \neq P_{te}(x,y)$ \citep{quionero2009dataset}. From this setting, the different dataset shifts can be characterized depending on the cause of the distribution change:
	\begin{enumerate}
		\item \textbf{Covariate shift:} this occurs in $\mathcal{X} \longrightarrow \mathcal{Y}$ domains when the independent variable distribution changes, that, is, $P(x)$ changes, but $P(y|x)$ remains constant.
		\item \textbf{Prior probability shift:} this occurs in $\mathcal{Y} \longrightarrow \mathcal{X}$ domains when the target variable distribution changes, that, is, $P(y)$ changes, but $P(x|y)$ does not.
		\item \textbf{Concept shift:} the relationship between independent variables and target variable changes. If this occurs in a $\mathcal{X} \longrightarrow \mathcal{Y}$ domain, then $P(y|x)$ changes but $P(x)$ does not. If it occurs in a $\mathcal{Y} \longrightarrow \mathcal{X}$ domain, then $P(x|y)$ changes but $P(y)$ remains constant.
	\end{enumerate}

This paper focuses on covariate shift. Hence, the point of departure is the assumption that i) the distribution of the covariates $x$ changes between training and test datasets, that is, $P_{tr}(x) \neq P_{te}(x)$ and that ii) the conditional distribution of the target remains constant, that is, $P_{tr}(y|x) = P_{te}(y|x)$.

One of the warhorses in covariate shift is the error estimation \citep{sugiyama2007covariate}. The real error in supervised learning is obtained by calculating how far off the prediction made is from reality. In convectional data sets, the actual error rate cannot be exactly computed since the actual values are not available. But several methods, like cross validation, are widely known and accepted to produce accurate error estimations. After the model is learned from a training data, it will yield an error similar in training and testing data. This assumption is made under the condition that $P_{tr}(x,y)$ and $P_{te}(x,y)$ remains constant.  The model performance is evaluated several times to ensure that the results are independent from the training data. In fact, the training dataset is split in folders where some of them simulate training data and others simulate test data. This is so, since one expects that these test folders represent, in one sense, the test data.  In covariate shift, this assumption is not held, since the distribution of test data may differ from that existing in training data. So, it may not possible to find suitable folds in training data to simulate test data. Consequently, such methods may lead to unreliable or inaccurate error estimations. In fact, it has been shown that using conventional error estimators is highly slanted in covariate shift contexts \citep{sugiyama2007covariate}. Under this paradigm, it is well known that robust error estimation must take into account the covariate distribution both in training and test datasets. An appealing procedure to cope with this drawback consists in adapting existing error estimation methods, as it is the case of cross validation, but including a correction factor in a attempt to make training data distribution, $P_{tr}(x)$, be as close as possible to the test data distribution, $P_{te}(x)$. This approach is called likelihood cross validation \citep{shimodaira2000improving} and works considering the importance concept as a correction factor. The importance is defined by the ratio between test and training covariate density functions, that is, if $p_{tr}(x)$ and $p_{te}(x)$ are the training and test covariate density functions, then, the importance is computed as:
\[
    w(x) = \frac{p_{te}(x)}{p_{tr}(x)}
\]

The idea is to compute the importance of each training example $x$. This means that the importance measures the ratio between the probability of a training example $x$ falling in the test dataset and the probability of this example falling in the training data set. Theoretically and taking into account this definition, the importance equals to $1$ if the dataset does not suffer from covariate shift. Once the importance of all training examples is computed, the example error is weighed by its importance, giving more weight to those errors committed on the examples that have a similar distribution to the examples in the test set.  This means that the error estimation will no longer be calculated as just the arithmetic average between all the errors for each example across the folds of cross validation. Instead, each estimated error over a training example $x$ obtained with cross validation, $ee(x)$, is weighted by multiplying it by its corresponding importance, $w(x)$, that is:
\[
iee(x)= ee(x) \cdot w(x)
\]

This way, the most relevant examples (the ones that are more similar to those in the test set) are more important towards the calculation of the error estimation, and thus, the effects of the covariate shift are alleviated. 

\begin{figure}[H]
	\centering
	\includegraphics[width=\textwidth]{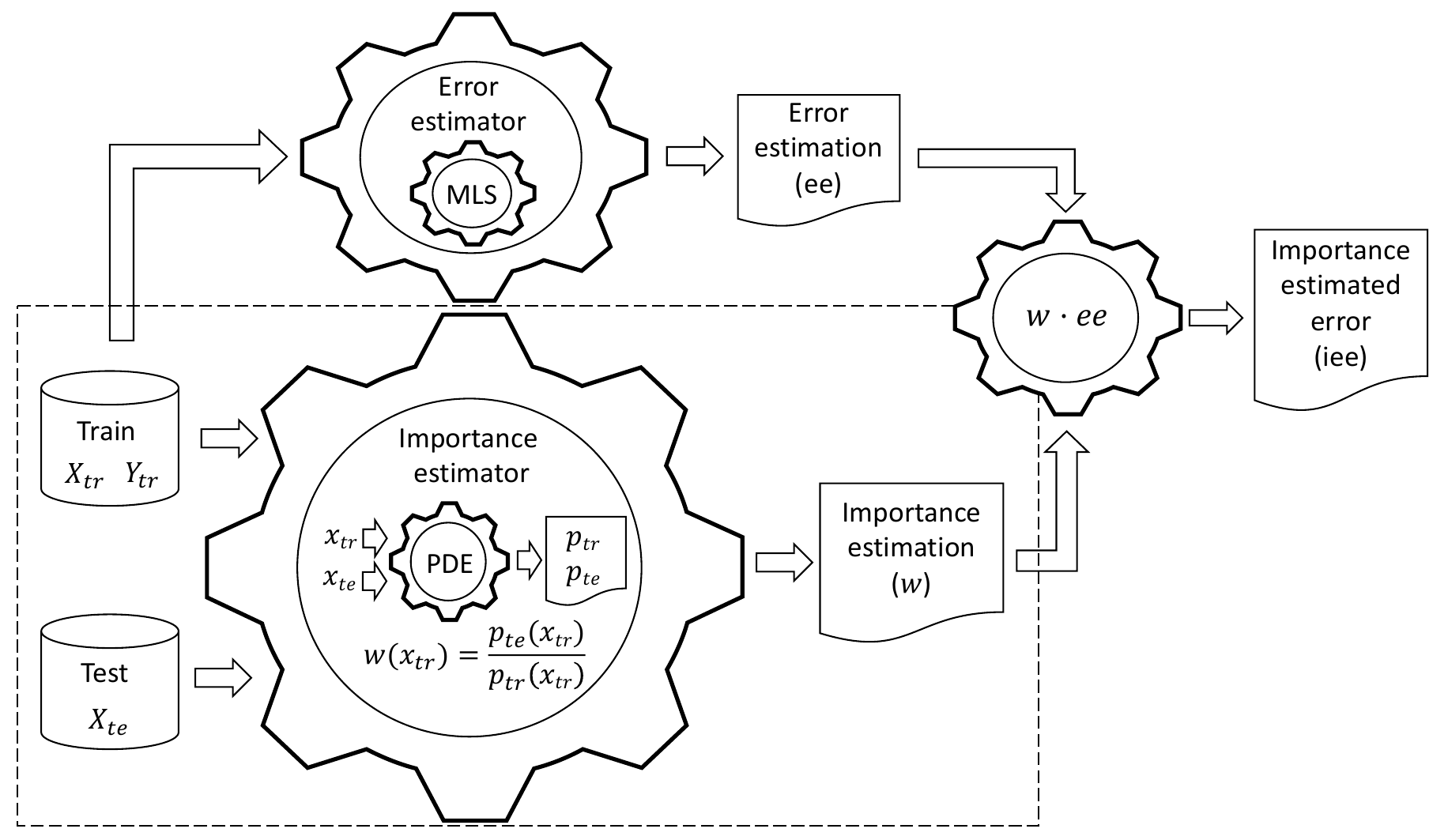}
	\caption{Importance estimated error under covariate shift. The importance estimated error is the product of the estimated error and the importance. The estimated error is computed from training data via a machine learning system (MLS). The importance is estimated from the covariates of both training and test data via a probabilistic density estimator (PDE).}
	\label{fig:importancediagram}
	\centering
\end{figure}

The hurdle now is that the importance is usually unknown and it is not a trivial task to estimate it. Therefore, providing promising importance estimations is now one of the main challenges in covariate shift.

Figure \ref{fig:importancediagram} summarizes the importance estimated error under covariate shift. From training data an error estimation $ee$ is computed using a machine learning system (MLS) (see top of Figure \ref{fig:importancediagram}). Also, for training an test covariates, the importance $w$ is computed using a probabilistic density estimator (PDE) (see the dashed rectangle of Figure \ref{fig:importancediagram}). Later on, a deep explanation of this computation will be detailed.

\section{Existing methods for importance estimation}\label{sec:existing}
Although covariate shift has been studied extensively in the literature, there not exist so many methods to estimate the importance under covariate shift. There are, however, some proposals to work under covariate shift which might be capable of doing this estimation in a more promising way.

Logistic regression (LR) \citep{bickel2007discriminative} is an approach that directly estimates the importance using a probabilistic classifier through a discriminative learning. Then, this method can be used to approximate $P(y|x)$ by discriminating test examples $\{x_i\}_{i=1}^{n_{te}}$ from training examples $\{x_j\}_{j=1}^{n_{tr}}$. Then, the training and testing examples will be respectively labelled as negative $y = -1$ and $y = 1$ positive examples. Hence,
\[
P_{tr}(x)=P(x| y = -1)
\]
\[
P_{te}(x)=P(x| y = 1)
\]

Now, applying the Bayes theorem,
\[
P_{tr}(x)=\frac{P(y = -1|x)}{P(y=-1)}\]
\[
P_{te}(x)=\frac{P(y = 1|x)}{P(y=1)}\]

Since the importance can be expressed in terms of the ratio of the distribution functions instead of the ratio of density functions, the expression will be:
\[
w(x) = \frac{P_{te}(x)}{P_{tr}(x)}=\frac{P(y = -1)\cdot P(y = 1|x)}{P(y = 1)\cdot P(y = -1|x)} \]

Moreover, $P(y = -1)$ and $P(y = 1)$ can be estimated by the number of training and test examples. Then, if $n_{tr}$ and $n_{te}$ are respectively the number of training and test examples, the importance will be computed as:
\[
w(x) = \frac{n_{tr}\cdot P(y = 1|x)}{n_{te}\cdot P(y = -1|x)}
\]

The main drawback of the discriminative approaches is that training may be time consuming.

The Kernel Mean Matching (KMM) \citep{gretton2009covariate} is another method for importance estimation. It has the property of avoiding density estimation $p(x)$, and consequently, it does need the covariate distribution $P(x)$ to be known. This method matches training and test sets in feature space defined by a kernel. The KMM process is non-parametric that performs a quadratic programming optimization in the kernel space that attempts to weight the training examples such the means of both training and test sets are as close as possible. That is, the basic idea of KMM is to find a $w(x)$ so that the maximum difference (called maximum mean discrepancy) in the kernel space of the means of $P_{tr}(x)\cdot w(x)$ and $P_{te}(x)$ is minimized.

One of the advantages of KMM is that, since it does not require density estimation, since the $w(x)$ is directly estimated. This property makes expected the method to work well in high-dimensional cases. However, the performance of KMM depends largely on the parameter tuning of the kernel that determines the feature space where the optimization takes place, which cannot be optimized in a simple way.

One of the downsides of KMM is that this method tends to be hindered by the quadratic complexity of calculating and sorting the kernel matrices over training and test data. To solve this issue, Ensemble Kernel Mean Matching (EKMM) \citep{miao2015ensemble} was proposed. This approach arbitrarily split the data test into a partition of size $p$, $\{S_k\}_{k=1}^{p}$, estimating the density ratios of each of partition component and then fusing these estimations using a weighted sum of the components of the partition. The weight is just the proportion size of the component with regard to the size of the original test dataset. That is, $w(x)$ is computed as
$$w(x)=\sum_{k=1}^{p} \frac {|S_k|}{n_{te}} \cdot w^{k}(x)$$
where $|S_k|$ is number of examples in the test data and $w^k(x)$ is its corresponding density ratio estimation. Some advantages of EKMM over KMM are its suitability for distributed implementation, a lower error bound and a higher accuracy.

One method prominent in the literature is the use of a Kernel Density Estimator (KDE) \citep{silverman1986density}.  This is a non-parametric technique to estimate a density $p(x)$ from a set of examples $\{x_k\}_{k=1}^n$.  Considering a kernel $K_\sigma$ (where $\sigma$ is a parameter that represents its width), then KDE is expressed as:

\[
p(x) = \frac{1}{n(2\pi\sigma^2 )^{d/2}} \sum_{k=1}^{n} K_\sigma(x-x_k)
\]

This density estimation is performed for both training examples and test examples and then, the importance is computed as the ratio. The estimation of KDE depends on the choice of the kernel bandwidth $\sigma$. However, a potential limitation of this approach is that KDE suffers from the \textit{curse of dimensionality} \citep{hardlenonparametric}, which means that the number of examples needed to maintain the same approximation quality grows exponentially as the dimension of the input space increases. Therefore, KDE may not be reliable in high-dimensional cases.

The most promising method for importance estimation nowadays is the Kullback-Leibler Importance Estimation Procedure Distribution (KLIEP) \citep{sugiyama2012machine}, which can be considered as an improvement of KDE. KLIEP performs an optimization consisting of minimizing the Kullback-Leibler divergence. The optimization problem is convex, so the unique global solution can be obtained. It is based off the density ratio estimation, which says that it is enough to know the ratio of the probability densities when there is a distribution change, even if the new distributions are unknown. Hence, it also avoids density estimation $p(x)$ and does not need covariate distribution $P(x)$ to be known, as KMM does. However, KLIEP minimizes the Kullback-Leibler divergence and obtains a linear function for the importance expressed by:
\[
	w(x) = \sum_{k=1}^{b} \alpha_k \phi_k (x),
\]

where $\{\phi_k (x)\}_{k=1}^b$ are basis functions such that $\phi_k(x) \geq 0$ for $k=1, \dots, b$ and $\{\alpha_k\}_{k=1}^b$ are parameters to be learnt by minimizing the Kullback-Leibler divergence from the estimated test density $\overline{p}_{te}(x) = w(x)\cdot p_{tr}(x)$ to the real test density $p_{te}(x)$. Basically, the optimization criterion of KLIEP is as follows:
\[
\underset{\{\alpha_k\}_{k=1}^b}{\text{maximize}} \sum_{j=1}^{n_{te}} \log \left( \sum_{k=1}^{b} \alpha_k \phi_k (x_j)\right)
\]
\[
\text{subject to } \sum_{i=1}^{n_{tr}} \sum_{k=1}^{b}\alpha_k \phi_k (x_i) = n_{tr} \text{ and } \alpha_1, \alpha_2, ..., \alpha_b  \geq 0
\]

An advantage of LR over KMM and KDE is that it uses a standard supervised classification problem to estimate the sample weights. Unfortunately, when the covariate shift is substantial and the number of examples in training is small, importance weighting often leads to very high variance estimates \citep{cortes2014domain} and to inaccurate models.

As KMM and KLIEP do not involve density estimation, they do not suffer from the curse of dimensionality in the same way that KDE does. However, and despite both KMM and KLIEP directly produce importance estimates instead of estimating densities, KLIEP is practically more useful. This is so because KLIEP presents the advantage of incorporating its own model selection to the procedure and thus, it avoids a separate optimization procedure to properly tune its parameters like KMM needs.

Although KLIEP seems to be the method to offer more promising performance, this paper explores the impact of employing target information in the importance computation of the above-detailed methods.

\section{Improving importance estimation by including target information} \label{sec:methods}
The most prevalent methods in the literature about estimating the importance under covariate shift are centered on the concept of importance as a weight vector to adapt the error estimation. All of them have in common the only use covariate information in order to estimate the importance, some of them by means of probability distributions and others by means of density estimations. Thus, they all ignore to include the target information in the importance computation. In this direction, this paper works under the hypothesis that target information would be a promising information source that would lead to a potential improvement in the performance of error estimation.

Until now and according to previous sections $w(x)$ was defined as the proportion of density estimation of test and training data just using the covariates, that is,
\[
    w(x) = \frac{p_{te}(x)}{p_{tr}(x)}
\]

Now and in order to generalize the definition of the importance to enrich the information, let us define a function $\phi$ over the feature space $\mathcal{X}$ that maps an example $x$ over a vector $v$ in the space $\mathcal{F}$ what will fed the density functions $p_{te}$ and $p_{tr}$, that is

\[
\begin{array}{lcll}
\phi:&\mathcal{X}&\rightarrow &\mathcal{F}\\
&x&\rightarrow &\phi(x)=v
\end{array}
\]

Hence, now the importance $w(x)$ is redefined in terms of the function $\phi$ as
\[
    w(x) = \frac{p_{te}(\phi(x))}{p_{tr}(\phi(x))}
\]
The feature space $\mathcal{F}$ can be of any dimension. Also, the function $\phi$ is expected to collect information of an example $x$. Let notice that this definition encapsulates the original definition of the importance equalizing the function $\phi$ to the identity function, that is, $\phi(x)=x$. In this particular case, $\mathcal{F}=\mathcal{X}$, then, the dimensions of the feature space and image space of $\phi$ are equal. 

This generalization of the importance makes possible to include a feature selection function $\phi$ in order to find the most informative subset of covariates, which may lead to an improvement in the error estimation performance. However, the interest of this paper lies in defining $\phi$ in a sense that is able to collect the relationship between the features $x$ and the targets $y$. A priori, the main drawback of this purpose is the unavailability of target information in testing for computing $p_{te}(\phi(x))$. However, a function $f:\mathcal{X}\rightarrow\mathcal{Y}$ can be learned from training covariates $\{x_i\}_{i=1}^{n_{tr}}$ and targets $\{y_j\}_{j=1}^{n_{tr}}$, which might be a candidate for reproducing the relationship between covariates and targets. The actual values of the targets in training can be taking for computing $p_{tr}(\phi(x))$. However, it does not seem a good practice. This is so because, it has not sense to consider actual values for training and predicting values for testing in a ratio computation as the importance is. Hence, the proposal is to work with predictions both in training for computing $p_{tr}(\phi(x))$ and in testing for computing $p_{te}(\phi(x))$. Particularly, the proposal consists of replacing the covariates by the predictions produced by $f$. This idea is quite sensible because it is reasonable to assume that the predictions summarize the covariate information with regard to the targets. One can think that it is not proper to apply $f$ to the testing dataset, since the distribution in training and testing differs. However, covariate shift assumes that $P(y/x)$ remains constant in training and testing, despite $P(x)$ changes, and, inducing $f$ means in fact to estimate $P(y/x)$. This means to propose directly $f(x)$ as $\phi$ function, that is, $\phi(x)=f(x)$ and hence $\mathcal{F}=\mathcal{Y}$. This proposal reduces the dimension of $\mathcal{F}$, from being the dimension of $\mathcal{X}$ to be the dimension of $\mathcal{Y}$, which is quite lower. Clearly, the predictions yield by $f$ represent the covariates, and, despite the estimation of $f$ would not be perfect, it summarizes in one sense the information contained in the covariates, hopefully removing the possible noise or redundancy contained in the covariates. 

Another proposal of $\phi$ would be to consider both information, that coming from the targets (again in form of predictions rather than in form of actual values) and that coming from the covariates, that is, $\phi(x)=(x,f(x))$. In this case, the dimension of $\mathcal{F}$ will be higher and equal to the sum of the dimensions of $\mathcal{X}$ and $\mathcal{Y}$, since $\mathcal{F}=\mathcal{X}\times \mathcal{Y}$. This proposal increases the dimension of $\mathcal{F}$, but, on one hand the use of covariates may compensate the imperfection of the $f$ estimation and, on the other hand, the use of the predictions yield by $f$ gives knowledge to the importance estimation in spite of the redundancy it may add.

\begin{figure}[H]
	\centering
	\includegraphics[width=\textwidth]{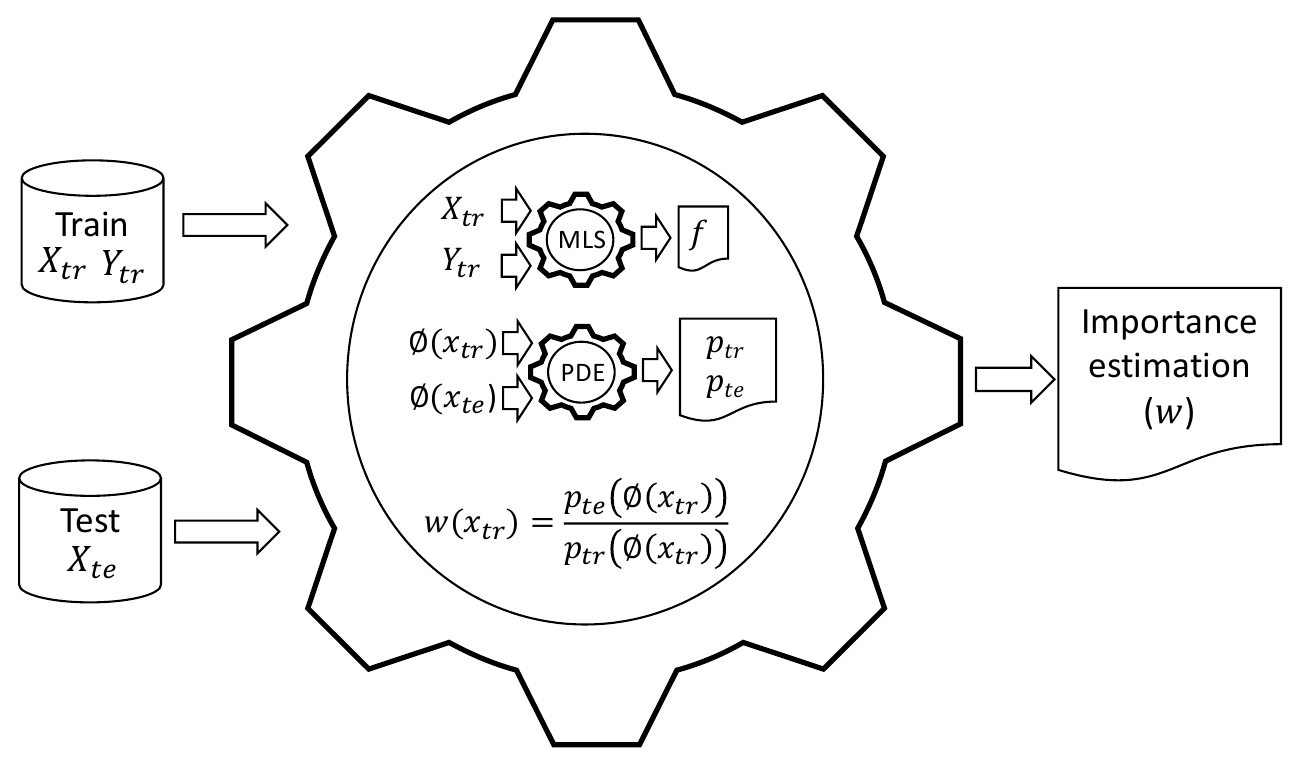}
	\caption{Summary of the importance computation using a $\phi$ function}
	\label{fig:importancediagramparticular}
	\centering
\end{figure}

Figure \ref{fig:importancediagramparticular} summarizes the way of computing the importance using a $\phi$ function. If $\phi$ involves predictions (as the proposed $\phi$ functions, namely, $\phi(x)=f(x)$ and $\phi(x)=(x,f(x))$), a MLS is needed to obtain a model $f$ from the training data ($X_{tr}$ and $Y_{tr}$)  in order to compute the predictions. This does not happen if $\phi$ equals the identity function ($\phi(x)=x$), which is the original way of computing the importance. The density functions $p_{tr}$ and $p_{te}$ are obtained using a PDE fed by both training and test covariates after being transformed using the $\phi$ function. Finally, the importance $w$ is computed as the ratio between $p_{te}$ and $p_{tr}$ applied over the transformation using $\phi$ of the training covariates.

\begin{figure}[H]
	\centering
	\includegraphics[width=\textwidth]{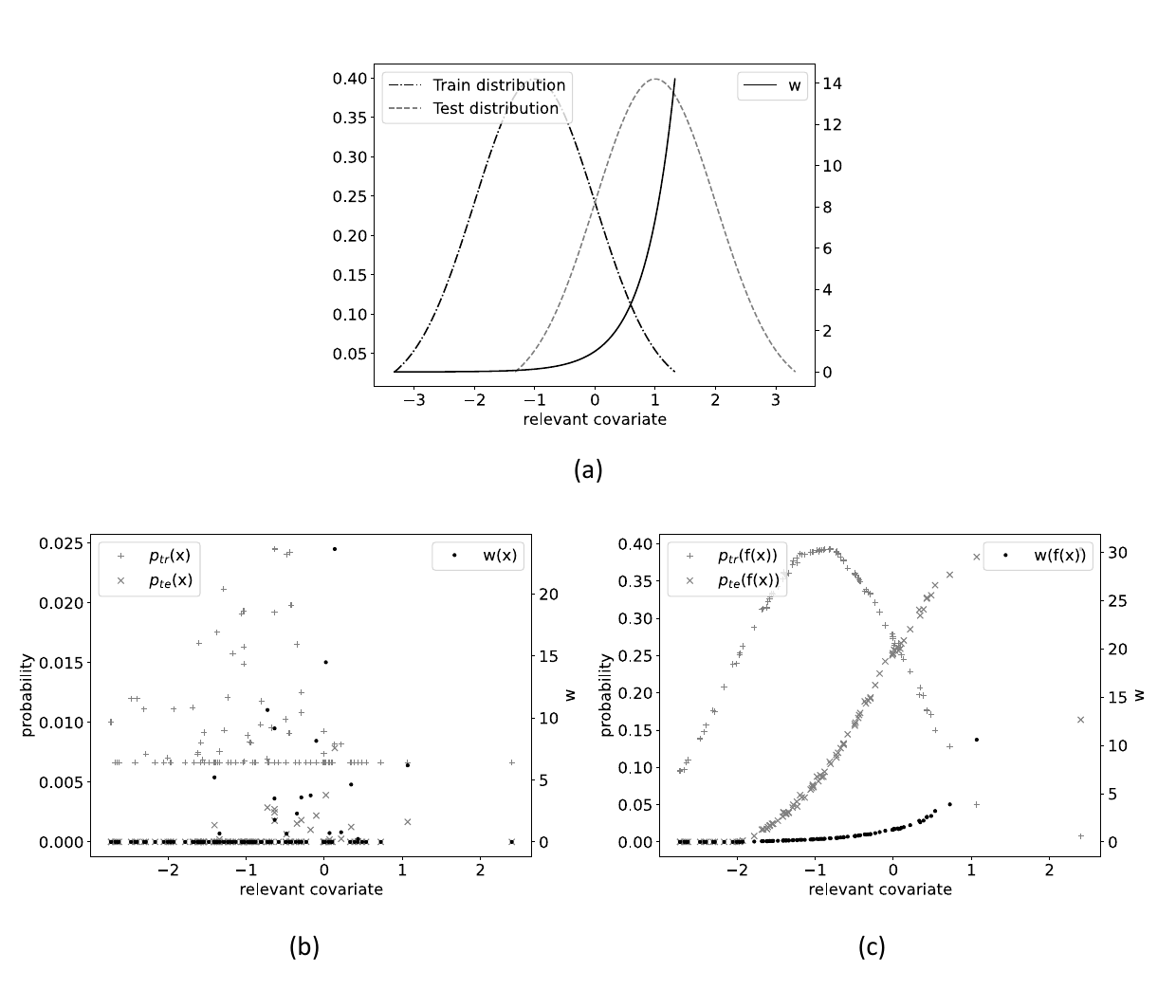}
	\caption{A toy example for motivating the use of the predictions in the computation of the importance}
	\label{fig:toyexample}
	\centering
\end{figure}

Figure \ref{fig:toyexample} displays a toy example to illustrate how using predictions for computing the importance is more in accordance to the distribution change between training and test data than the original way of computing the importance. The toy example consists of $100$ training instances and $100$ test instances with five covariates, but just one is relevant for the target. In Figure \ref{fig:toyexample} (a) the relevant covariate in training and test data follows a normal distribution, but with the average of test data shifted to the right with regard to the training data. The theoretical importance $w$ is shown in solid line. Figures \ref{fig:toyexample} (b) and (c) display the estimated probability of the training instances of following the training distribution $p_{tr}$ and of following the testing distribution $p_{te}$. They also shown the estimated importance ($w$). The difference between (b) and (c) is that in the former $p_{tr}$ and $p_{te}$ are computed just using the covariates ($\phi(x)=x$), whereas in the latter $p_{tr}$ and $p_{te}$ are computed using the predictions yield by the $f$ model ($\phi(x)=f(x)$). As it can be seen, estimating the importance using the predictions clearly fits better the real situation displayed in Figure \ref{fig:toyexample} (a).

Before continuing, let us clarify that the MLS taken for inducing $f$ to compute the predictions in order to estimate the importance can be independent and different  from the one that will be used to estimate the error, which will be weighed by the importance.

Let us now establish the differences of applying this procedure in classification and regression. In classification, a model $f$ produces a discrete value. These values do not seem adequate to feed the density functions because it loses quite valuable information. In order to collect the most information contained in the covariates through the model $f$, we propose to focus on those classification approaches that produce a numerical value before a threshold is applied to give the discrete values of the classification.  This numerical value clearly contains quite more information of the covariates that the discrete yielded afterwards. Hence, the proposal for classification is to set $\phi(x)=f^*(x)$ instead of $f(x)$. This adjustment in classification task it is not necessary for regression. Let us notice that in case of multiclass classification, the one-versus-the-rest procedure is applied and as many values as number of classes are produced. In general, any Error Correcting Output Codes (ECOC) (that encapsulates the one-versus-the-rest procedure) is also valid.

Under the new general definition of the importance $w(x)$ though a function $\phi(x)$, all LR, KMM, EKMM, KDE and KLIEP methods detailed in Section \ref{sec:existing} will be adapted to this generalization.

\section{Experiments}
This section deals with the experiments carried out to check the impact of including target information in the importance estimation performed by some existing procedures for both classification and regression predictions.

Firstly, the lack of a benchmark dataset in which the changes are known makes arise the need of preparing a series of datasets of this kind. Section \ref{sec:datasetgeneration} details the procedure carried out for this purpose. Plankton classification real application that suffers from covariate shift and motivates this research is described in Section \ref{sec:realproblem}. Then, experimental settings are shown in Section \ref{sec:settings}. Finally, the experimental results are discussed in Section \ref{sec:results}.

\subsection{Including controlled distribution changes in datasets}\label{sec:datasetgeneration}
This section describes the procedure carried out to include controlled distribution changes in datasets. Most of the datasets were taken from the UCI Machine Learning Repository\footnote{\url{https://archive.ics.uci.edu/ml/index.php}}, except titanic dataset that was taken from Stanford University \footnote{\url{https://web.stanford.edu/class/archive/cs/cs109/cs109.1166/problem12.html}}. Just changes in $P(x)$ distribution are needed in order to fit the covariate shift criteria. The sampling procedure to do so will be based on an already exiting idea in the literature \citep{perez2017using}. This procedure only works for binary classification; hence, a multiclass adaptation must be tackled. The existing procedure for binary classification first consists of randomly selecting the sample prevalence $p$ in $[0..1]$ for the positive class. Then, simple random sampling with replacement is performed within the positive class examples until the number of positive examples, given by the chosen prevalence $p$, is obtained. This process guarantees $P(x|y)$ to be constant. The same operation is repeated for the negative class, with its prevalence being equal to $n = 1 - p$. Then, the desired diversity is obtained by changing the prevalence of each generated sample.

This paper extends that basic idea to more types of learning than just binary classification. For this purpose, our proposal slightly modifies the random sampling with replacement procedure to fit multiclass context. Also, a more sophisticated modification must take place in order to cope properly with regression. The next sections respectively detail the modifications for multiclass and regression paradigms.

\subsubsection{Procedure for classification problems}
The idea is to divide datasets in training and set sets, but making sure that the test sets have a different distribution in the covariates $x$ than the training set. To do this, a standard split will divide the examples into training and test sets, fixing certain percentage of test examples. The training set will be unchanged, but the test set will be taken as base to draw the examples for the different distributions sets. Then, given a random seed, the procedure will create one training set and several test sets with different distributions in the covariates $x$. The process to extract a new distribution from the base test set is simple. We create an uniform multivariate distribution with a number of variables equal to the number of classes, and with a minimum distance $d$ between the probabilities, corresponding to the minimum prevalence $p$ that will be assigned to the classes. The prevalence $p$ values for each class were randomly chosen in the interval $[5 − 95\%]$ \citep{perez2017using} in order to deliberately avoid values near $0\%$ and $100\%$.  The reason of this decision is because some difficulties may arise at these points due to the lack of examples in some of the classes. Given the selected prevalence $p$ for certain class, random sampling with replacement (to ensure $P(x|y)$ remains constant) was used to generate the number of examples required for each class. D'Hondt method \citep{d1882systeme} is applied to generate a discrete distribution for all the classes from the multivariate probability distribution. Table \ref{tab:classification-datasets} shows the number of instances, covariates and classes for the datasets obtained.

\begin{table}[h!]
\centering
\begin{tabular}{@{}lrrr@{}}
\toprule
\textit{Dataset}     & Instances & Covariates & Classes \\ \midrule
  {iris}        &   150        &    4        &    3     \\
  {sonar}       &   208        &     60       &    2     \\
  {ionosphere}  &   351        &   34         &     2    \\
  {cmc}         &  1473         &   9         &    3     \\
  {habermans}   &  306         &   3         &    2     \\
  {transfusion} &  748         &   5         &    2     \\
  {wdbc}        &  569         &   32         &    2     \\
  {spectf}      &  267         &    44        &    2     \\
  {splice}     &  3175         &    61        &   3      \\
  {titanic}     &  2201         &    4        &   4      \\ \bottomrule
\end{tabular}
\caption{Number of instances, covariates and classes for classification datasets.}
\label{tab:classification-datasets}
\end{table}

\subsubsection{Procedure for regression problems}
In order to ensure significant distribution changes between training and test examples in regression, those examples with low target values may be taken as training examples, whereas those examples with higher target values may be taken as a test (or vice-versa). Using a threshold may be a simple way to establish what is considered as low and as high.  The main drawback of this proposal is that no target value overlapping occurs between training and testing examples, hence, a good learning is almost impossible. In order to avoid this situation, the proposal is to smooth the procedure using a probabilistic split. In this direction, the idea performed consists of establishing that low (or high) value target examples will have more probability of falling into the training set and conversely, high (or low) value target examples will have more probability of falling into the test set.

First of all, the target values are normalized in the interval $[-1;+1]$. Then, a sigmoid density probability function is taken as probabilistic split. This function is expressed as
\[
\frac{1}{(1+e^{(-\gamma \cdot x)})}
\]
where $\gamma$ is a parameter that conditioned the split, that is, i) the higher the value of $\gamma$, the higher the target value differences between training and test examples will be (and the learning will be almost impossible) ii) conversely, the lower the value of $\gamma$, the lower the target value differences between training and test examples will be (and the data shift situation will almost disappear). Hence, the ideal situation will be to chose a moderate value of $\gamma$ to remain the covariate shift, but without falling in a almost-impossible learning task. Table \ref{tab:regression-datasets} shows the number of instances, covariates and classes for the datasets obtained.

\begin{table}[h!]
\centering
\begin{tabular}{@{}lrr@{}}
\toprule
\textit{Dataset}               & Instances & Covariates \\ \midrule
abalone               &   4177       &    8                 \\
computer-hardware   &   209        &    9                 \\
{wine-quality-red}      &    1599      &    12                 \\
{wine-quality-white}    &    4898      &    12                \\
{auto-mpg}              &    398       &    8              \\
{autos}                 &    205       &    26             \\
{student-mat}           &    395       &    32           \\
{student-por}           &    395       &    32         \\
{residential-v9}        &    373       &    28             \\
{residential-v10}       &    373       &    28            \\
{ticdata}               &    5822      &    43                \\
\bottomrule
\end{tabular}
\caption{Number of instances and covariates for regression datasets.}
\label{tab:regression-datasets}
\end{table}

\subsection{Real world problem application}\label{sec:realproblem}
A real world application under covariate shift that gives origin to this research is plankton classification. Plankton plays a vital role in marine ecosystems, then, performing good predictions is a clue task in order to know the kind of species that live in certain ocean areas. The main pitfall arisen in this context is that the characteristics and life cycles of the different plankton classes change throughout the year, which makes appear certain species in areas where they have not been seen before. Thus, the distribution of the classes changes depending on when the examples are drawn. Obviously, this situation fits covariate shift due to the distribution of the covariates $P(x)$ changing between training and test sets, but the conditional target distribution $P(y|x)$ remaining unchanged.

The WHOI-Plankton dataset \citep{orenstein2015whoi} has been used for the experimentation. In particular, since the original dataset is a computer vision problem, the deep features obtained by a CNN as found in \citep{10.1093/plankt/fbz023} were taken as covariates, whereas the original plankton classes were maintained, which leads to a high-dimensional classification task under covariate shift.

As stated before, the class distribution changes depending on when the examples are drawn. Thus, two sets with changes in the covariate distributions may be taken splitting the examples drawn during the first half of the year apart from the examples drawn during the second half of the year. The dataset has examples from $7$ different years, from 2006 to 2013.  The original dataset has 51 classes, but those examples from the classes labelled "bad", "na" and those from the functional group "other" were discarded. All datasets have 514 covariates, which are the deep features extracted from the original images.  The number of instances for each yearly dataset can be found in Table \ref{tab:plankton-datasets}.

\begin{table}[h!]
	\centering
	\begin{tabular}{@{}lr@{}}
		\toprule
		\textit{Dataset}               & Instances \\ \midrule
		p-2006          &  20552 \\
		p-2007          &  40424 \\
		p-2008          &   43285\\
		p-2009          & 48131  \\
		p-2010          & 59427  \\
		p-2011          & 46716  \\
		p-2012          & 43406 \\
		p-2013           & 53074 \\\bottomrule
	\end{tabular}
	\caption{Number of instances for the plankton yearly datasets.}
	\label{tab:plankton-datasets}
\end{table}

\subsection{Experimental settings} \label{sec:settings}
This section presents the experimental settings. Concerning the parameters for the distribution changes in the covariates $x$ to simulate covariate shift in the artificial datasets, $20$ different test data from base test dataset were generated per each of the $5$ seeds (in the range $[2032, 2036]$), leading to a total of $100$ different test set with covariate shift, containing $33\%$ of the examples on the test sets. In classification, the distance $d$ for establishing differences in the probabilities were set to $1 / (10 \cdot m)$ (remember that $m$ is the number of classes). In regression, the parameter  $\gamma$ of the sigmoid function were fixed to $5$ and $-5$. The election of this value is motivated for the fact of avoiding a sigmoid either similar to a step function or linear function.

In case of the plankton dataset, once divided in halves of the year, the training and test sets are paired for the experimentation in 15 different combinations of training and test sets were obtained taking into account the curse of the time. This means that the training examples of the first dataset belongs to the first half of the year 2006 and the test examples to the second half of the year 2006. Then, the training examples of the second dataset belongs to the second half of the year 2006 and the test examples to the first half of the year 2007 and so on. Then, let us label the first half of the year dataset with the subindex $a$ and the second half of the year dataset with the subindex $b$. Hence, the $15$ datasets combinations will be p2006a-2006b, p2006b-2007a, p2007a-2007b, p2007b-2008a, p2008a-2008b, p2008b-2009a, p2009a-2009b, p2009b-2010a, p2010a-2010b, p2010b-2011a, p2011a-2011b, p2011b-2012a, p2012a-2012b, p2012b-2013a and p2013a-2013b. Table \ref{tab:placktonexperiments} shows the number of the instances for each of the $15$ combinations.

\begin{table}[h!]
	\centering
	\begin{tabular}{@{}lrllr@{}}
		\toprule
		\textit{Dataset} & Instances && \textit{Dataset} & Instances \\ \midrule
		p2006a-2006b     & 20552    & & p2010a-2010b     & 59427     \\
		p2006b-2007a     & 45270    & & p2010b-2011a     & 41629     \\
		p2007a-2007b     & 40424    &  & p2011a-2011b     & 46716     \\
		p2007b-2008a     & 27387    & & p2011b-2012a     & 46003     \\
		p2008a-2008b     & 43285   &  & p2012a-2012b     & 43406     \\
		p2008b-2009a     & 48313    & & p2012b-2013a     & 37407     \\
		p2009a-2009b     & 48131    & & p2013a-2013b     & 53074     \\
		p2009b-2010a     & 69320    & &                  &           \\ \bottomrule
	\end{tabular}
	\caption{Number of instances for the $15$ experimentation combinations of the plankton datasets.}
	\label{tab:placktonexperiments}
\end{table}

The performance measures to assess the quality of the importance estimators will be the distance between the actual error and the estimated error. In fact, the goal is not to check how good the predictions are, but just how far off the error estimations are from the actual value. For this purpose, a 10-fold log likelihood cross validation will be employed to compute the estimated error. Then, LR, KMM, EKMM, KDE and KLIEP are compared when they use covariates (LR-C, KMM-C, EKMM-C, KDE-C and KLIEP-C), when they use predictions (LR-P, KMM-P, EKMM-P, KDE-P and KLIEP-P) and when they use both covariates and predictions (LR-CP, KMM-CP, EKMM-CP, KDE-CP and KLIEP-CP). Furthermore, a Friedman test \citep{demvsar2006statistical,garcia2008extension, pacifico2018plant} to the distances was performed in order to make the comparison easier, since this test  obtains a ranking of the performance of the methods. This test consists of a comparison in two steps for each measure. The first step is a Friedman test that rejects the null hypothesis that states that not all learners perform equally. The second step is the post-hoc Nemenyi test. In the tables, the ranks of each dataset are shown in brackets. In case of ties, average ranks are indicated. The average ranks of overall datasets are calculated and presented in the last row of each table. The comparisons performed were done so the methods using covariates (-C) are compared against their counterparts using predictions (-P) and both covariates and predictions (-CP). This is so because our focus is on whether using target information (predictions made by a model) instead of covariate information or whether using both kinds of information (covariate and predictions) improves the importance estimation and leads to increase the error estimation accuracy. Hence, the comparison will be first LR-C, LR-P and LR-CP, then KMM-C, KMM-P and KMM-CP, then EKMM-C, EKMM-P and EKMM-CP, then KDE-C, KDE-P and KDE-CP, and lastly KLIEP-C, KLIEP-P and KLIEP-CP.

The experimental analysis is done using an implementation written in Python together with the \href{https://scikit-learn.org/stable/}{\texttt{scikit-learn library}}. The code can be found in this  \href{https://github.com/SaraTomillo/DCiML}{\texttt{GitHub repository}.} Concerning the parameters of the different methods, the decision of their values was according to the authors of the systems. Concerning KMM, Gaussian kernel was taken with bandwidth $\sigma$ equal to $1$, $b=1000$ and $\epsilon=(\sqrt{n_{tr}}-1)/\sqrt{n_{tr}}$. The parameters for EKMM were equal to those of KMM, adding the parameter of the partition $p$ that was fixed to $20$. The original version of EKMM only partitions the test data across the members of the ensemble while transferring all the training data to each partition component estimator. This is based on the assumption that the number of training data is usually small. Our proposal consists of altering this original approach by performing the ensemble sampling over the training data instead, given that the datasets used in this paper have a greater number of samples on the training data than in test. In regard to KDE, among the amount of kernels available in the literature for density estimation, Epanechnikov kernel has been shown to be more promising among the non-parametric kernels \citep{li2007nonparametric, samiuddin1990nonparametric}. The kernel bandwidth $\sigma$ taken was the default value of \texttt{sklearn}, which is $1$. In relation to KLIEP, a Gaussian kernel was chosen and the bandwidth $\sigma$ was optimized using a 3-folds cross validation over the values $\{0.01, 0.1, 0.25, 0.5, 0.75, 1\}$, according to authors. The bandwidth of the kernels acts as a smoothing parameter, controlling the trade-off between bias and variance in the result. However, let remember that in KMM and KDE, it was fixed to so using a value of $1$ again according to authors, since it guarantees that the density distribution will have neither a high bias nor a high variance.

Regarding the bandwidth of the kernels, their tuning may improve the error estimation performance, it would not have much effect, since the goal of the paper is not to improve the error estimation itself, otherwise it is to check whether the target information in the importance computation improves the performance with regard to consider covariate information or both.

Among the existing prediction methods, Support Vector Machine\footnote{\url{https://scikit-learn.org/stable/modules/generated/sklearn.svm.SVR.html}} \citep{cortes1995support} (SVM) and Support Vector Regression\footnote{\url{https://scikit-learn.org/stable/modules/generated/sklearn.svm.SVC.html}}(SVR) have been chosen to obtain the predictions respectively in classification and regression. Also, more experiments with Logistic Regression \footnote{\url{https://scikit-learn.org/stable/modules/generated/sklearn.linear_model.LogisticRegression.html}} (LogReg) for classification and Ridge Regression\footnote{\url{https://scikit-learn.org/stable/modules/generated/sklearn.linear_model.Ridge.html}} (Ridge) for regression were performed.

Despite it is possible to use different prediction methods to compute the importance and to estimate the error, the experiments were carried out using the same method for both tasks due to not present an excessive number of tables.



\subsection{Results discussion} \label{sec:results}
This section presents and discusses the results obtained. Let us first comment on the results for the artificial datasets and leave the correspondence ones for plankton dataset later. Tables \ref{tab:classification-SVM} and \ref{tab:classification-LR} expose the results for classification datasets when SVM and LogReg are taken for both estimating predictions and error. Also, Tables \ref{tab:regression-SVM}  and \ref{tab:regression-LR} display the results for regression datasets when SVR and Ridge are taken for both estimating predictions and error. As stated before, the focus of this research is not on how good the predictions are, otherwise on how accurate the error estimation is and if the accuracy increases when target information is taken into account in the importance computation. For this reason, the tables show the Friedman-Nemenyi test that compare the approaches one with each other (using just covariate information, using just target information or using both kings of information) rather than the distance error estimation. Then, the scores are the Friedman rankings, organized on triplets (three approaches for each method, as stated before).

\setlength\tabcolsep{4pt}
\begin{table}[!ht]
	\small
	\centering
	\begin{tabular}{@{}lrrrlrrrlrrrl@{}}
		\toprule
		\textbf{Dataset} & \multicolumn{4}{c}{\textbf{LR}}             & \multicolumn{4}{c}{\textbf{KMM}}                                & \multicolumn{4}{c}{\textbf{EKMM}}                                                                       \\ \cmidrule(l){2-13}
		& \textbf{-C} & \textbf{-P} & \textbf{-CP} &  & \textbf{-C} & \textbf{-P} & \textbf{-CP} &                      & \multicolumn{1}{l}{\textbf{-C}} & \multicolumn{1}{l}{\textbf{-P}} & \multicolumn{1}{l}{\textbf{-CP}}  &  \\ \midrule
		cmc                               & \textbf{1.97}                   & 2.06                            & \textbf{1.97}                    &  & \textbf{1.73} & 2.54          & \textbf{1.73} &                      & \multicolumn{1}{l}{\textbf{1.88}} & \multicolumn{1}{l}{2.03}          & \multicolumn{1}{l}{2.09}          &  \\
		haberman                          & 2.44                            & \textbf{1.12}                   & 2.44                             &  & 2.26          & \textbf{1.48} & 2.26          &                      & \multicolumn{1}{l}{1.98}          & \multicolumn{1}{l}{\textbf{1.96}} & \multicolumn{1}{l}{2.06}          &  \\
		ionosphere                        & 2.39                            & \textbf{1.22}                   & 2.39                             &  & \textbf{1.76} & 2.48          & \textbf{1.76} &                      & \multicolumn{1}{l}{\textbf{1.90}} & \multicolumn{1}{l}{2.09}          & \multicolumn{1}{l}{2.01}          &  \\
		iris                              & \textbf{2.05}                   & 1.90                            & \textbf{2.05}                    &  & \textbf{1.72} & 2.55          & \textbf{1.72} &                      & \multicolumn{1}{l}{2.22}          & \multicolumn{1}{l}{1.92}          & \multicolumn{1}{l}{\textbf{1.86}} &  \\
		sonar                             & 2.16                            & \textbf{1.68}                   & 2.16                             &  & \textbf{1.66} & 2.68          & \textbf{1.66} &                      & \multicolumn{1}{l}{1.98}          & \multicolumn{1}{l}{2.12}          & \multicolumn{1}{l}{\textbf{1.90}} &  \\
		spect                             & \textbf{1.84}                   & 2.32                            & \textbf{1.84}                    &  & \textbf{1.71} & 2.58          & \textbf{1.71} &                      & \multicolumn{1}{l}{\textbf{1.93}} & \multicolumn{1}{l}{\textbf{1.93}} & \multicolumn{1}{l}{2.14}          &  \\
		splice                            & \textbf{1.73}                   & 2.54                            & \textbf{1.73}                    &  & \textbf{1.73} & 2.54          & \textbf{1.73} &                      & \multicolumn{1}{l}{\textbf{1.85}} & \multicolumn{1}{l}{2.19}          & \multicolumn{1}{l}{1.96}          &  \\
		titanic                           & \textbf{1.81}                   & 2.38                            & \textbf{1.81}                    &  & \textbf{1.80} & 2.40          & \textbf{1.80} &                      & \multicolumn{1}{l}{2.14}          & \multicolumn{1}{l}{\textbf{1.90}} & \multicolumn{1}{l}{1.96}          &  \\
		transfusion                       & \textbf{1.22}                   & 1.82                            & 2.96                             &  & \textbf{1.54} & 1.84          & 2.62          &                      & \multicolumn{1}{l}{2.12}          & \multicolumn{1}{l}{2.05}          & \multicolumn{1}{l}{\textbf{1.83}} &  \\
		wdbc                              & 2.35                            & \textbf{1.30}                   & 2.35                             &  & \textbf{1.81} & 2.38          & \textbf{1.81} &                      & \multicolumn{1}{l}{2.02}          & \multicolumn{1}{l}{2.07}          & \multicolumn{1}{l}{\textbf{1.91}} &  \\ \midrule
		\textit{avg. rank}                & 2.00                            & \textbf{1.83}                   & 2.17                             &  & \textbf{1.77} & 2.35          & 1.88          &                      & \multicolumn{1}{l}{2.00}          & \multicolumn{1}{l}{2.03}          & \multicolumn{1}{l}{\textbf{1.97}} &  \\ \midrule\midrule
		\textbf{Dataset} & \multicolumn{4}{c}{\textbf{KDE}}             & \multicolumn{4}{c}{\textbf{KLIEP}}                                & \multicolumn{4}{c}{}                                                                       \\ \cmidrule(l){2-9}
		& \textbf{-C} & \textbf{-P} & \textbf{-CP} &  & \textbf{-C} & \textbf{-P} & \textbf{-CP} &                      & \multicolumn{1}{l}{} & \multicolumn{1}{l}{} & \multicolumn{1}{l}{}  &  \\ \cmidrule{1-9}
		cmc                               & \textbf{1.92}                   & 2.16                            & \textbf{1.92}                    &  & \textbf{1.98} & 2.04          & \textbf{1.98} &        &                                   &                                   &                                   &  \\
		haberman                          & \textbf{1.50}                   & 3.00                            & \textbf{1.50}                    &  & \textbf{1.50} & 3.00          & \textbf{1.50} &        &                                   &                                   &                                   &  \\
		ionosphere                        & \textbf{1.78}                   & 2.44                            & \textbf{1.78}                    &  & \textbf{1.76} & 2.48          & \textbf{1.76} &        &                                   &                                   &                                   &  \\
		iris                              & \textbf{1.97}                   & 2.06                            & \textbf{1.97}                    &  & \textbf{1.95} & 2.10          & \textbf{1.95} &        &                                   &                                   &                                   &  \\
		sonar                             & \textbf{1.92}                   & 2.16                            & \textbf{1.92}                    &  & \textbf{1.96} & 2.08          & \textbf{1.96} &        &                                   &                                   &                                   &  \\
		spect                             & \textbf{1.91}                   & 2.18                            & \textbf{1.91}                    &  & \textbf{1.94} & 2.12          & \textbf{1.94} &        &                                   &                                   &                                   &  \\
		splice                            & \textbf{1.76}                   & 2.48                            & \textbf{1.76}                    &  & \textbf{1.92} & 2.17          & \textbf{1.92} &        &                                   &                                   &                                   &  \\
		titanic                           & 2.36                            & \textbf{1.28}                   & 2.36                             &  & 2.36          & \textbf{1.28} & 2.36          &        &                                   &                                   &                                   &  \\
		transfusion                       & 2.87                            & 2.13                            & \textbf{1.00}                    &  & 2.80          & 2.19          & \textbf{1.01} &        &                                   &                                   &                                   &  \\
		wdbc                              & \textbf{1.94}                   & 2.12                            & \textbf{1.94}                    &  & \textbf{1.94} & 2.12          & \textbf{1.94} &        &                                   &                                   &                                   &  \\ \cmidrule{1-9}
		\textit{avg. rank}                & 1.99                            & 2.20                            & \textbf{1.81}                    &  & 2.01          & 2.16          & \textbf{1.83} &        &                                   &                                   &                                   &  \\ \cmidrule{1-9}
	\end{tabular}
	\caption{Friedman ranking on \textbf{classification datasets} using a \textbf{SVM} model for -P.}
	\label{tab:classification-SVM}
\end{table}

\begin{table}[h!]
	\small
	\centering
	\begin{tabular}{@{}lrrrlrrrlrrrl@{}}
		\toprule
		\textbf{Dataset} & \multicolumn{4}{c}{\textbf{LR}}             & \multicolumn{4}{c}{\textbf{KMM}}                                & \multicolumn{4}{c}{\textbf{EKMM}}                                                                       \\ \cmidrule(l){2-13}
		& \textbf{-C} & \textbf{-P} & \textbf{-CP} &  & \textbf{-C} & \textbf{-P} & \textbf{-CP} &                      & \multicolumn{1}{l}{\textbf{-C}} & \multicolumn{1}{l}{\textbf{-P}} & \multicolumn{1}{l}{\textbf{-CP}}  &  \\ \midrule
		cmc                               & 1.70                            & 2.20                            & \textbf{2.10}                    &  & \textbf{1.45} & 1.70          & 2.85          &                      & 2.85                 & 1.70                 & \textbf{1.45}        &  \\
		haberman                          & \textbf{1.25}                   & 1.75                            & 3.00                             &  & 2.10          & \textbf{1.65} & \textbf{2.25} &                      & 2.70                 & 2.25                 & \textbf{1.05}        &  \\
		ionosphere                        & \textbf{1.40}                   & 1.65                            & 2.95                             &  & \textbf{1.60} & 2.80          & \textbf{1.60} &                      & 2.25                 & 2.50                 & \textbf{1.25}        &  \\
		iris                              & 2.10                            & 2.40                            & \textbf{1.50}                    &  & 2.45          & 2.15          & \textbf{1.40} &                      & 2.20                 & \textbf{1.60}        & 2.20                 &  \\
		sonar                             & 2.35                            & \textbf{1.40}                   & 2.25                             &  & \textbf{1.30} & 2.85          & 1.85          &                      & 2.15                 & \textbf{1.45}        & 2.40                 &  \\
		spectf                            & 2.55                            & 1.85                            & \textbf{1.60}                    &  & 1.80          & 2.80          & \textbf{1.40} &                      & 2.00                 & 2.60                 & \textbf{1.40}        &  \\
		splice                            & 2.47                            & 1.95                            & \textbf{1.58}                    &  & 1.79          & 2.95          & \textbf{1.26} &                      & 1.68                 & \textbf{1.53}        & 2.79                 &  \\
		titanic                           & 2.05                            & 2.60                            & \textbf{1.35}                    &  & \textbf{1.70} & 2.55          & 1.75 &                      & 2.05                 & \textbf{1.35}        & 2.60                 &  \\
		transfusion                       & \textbf{1.05}                   & 1.95                            & 3.00                             &  & 2.65 & 1.70          & \textbf{1.65}         &                      & 2.70                 & 2.25                 & \textbf{1.05}        &  \\
		wdbc                              & \textbf{1.55}                   & 1.65                            & 2.80                             &  & 2.65          & 1.95          & \textbf{1.40} &                      & 2.55                 & \textbf{1.55}        & 1.90                 &  \\ \midrule
		\textit{avg. rank}                & \textbf{1.85}                   & 1.94                            & 2.21                             &  & 1.95 & 2.31 & \textbf{1.74} &                      & 2.31                 & 1.88                 & \textbf{1.81}        &  \\ \midrule \midrule
		\textbf{Dataset} & \multicolumn{4}{c}{\textbf{KDE}}             & \multicolumn{4}{c}{\textbf{KLIEP}}                                & \multicolumn{4}{c}{}                                                                       \\ \cmidrule(l){2-9}
		& \textbf{-C} & \textbf{-P} & \textbf{-CP} &  & \textbf{-C} & \textbf{-P} & \textbf{-CP} &                      & \multicolumn{1}{l}{} & \multicolumn{1}{l}{} & \multicolumn{1}{l}{}  &  \\ \cmidrule{1-9}
		cmc                               & 2.95                            & \textbf{1.15}                   & 1.90                             &  & 2.80          & 1.85          & \textbf{1.35} &                      &                      &                      &                      &  \\
		haberman                          & \textbf{1.35}                   & 1.95                            & 2.70                             &  & 2.90          & 2.10          & \textbf{1.00} &        &        &        &        &  \\
		ionosphere                        & \textbf{1.40}                   & 2.60                            & 2.00                             &  & 2.75          & 2.00          & \textbf{1.25} &        &        &        &        &  \\
		iris                              & \textbf{1.70}                   & 2.35                            & 1.95                             &  & 2.00          & \textbf{1.85} & 2.15          &        &        &        &        &  \\
		sonar                             & \textbf{1.50}                   & 2.35                            & 2.15                             &  & 2.15          & \textbf{1.55} & 2.30          &        &        &        &        &  \\
		spectf                            & \textbf{1.35}                   & 1.85                            & 2.80                             &  & 2.25          & 2.45          & \textbf{1.30} &                      &                      &                      &                      &  \\
		splice                            & 1.95                            & \textbf{1.79}                   & 2.26                             &  & 3.00          & 1.63          & \textbf{1.37} &        &        &        &        &  \\
		titanic                           & 2.10                            & 2.15                            & \textbf{1.75}                    &  & 2.15          & \textbf{1.20} & 2.65          &        &        &        &        &  \\
		transfusion                       & 2.20                            & \textbf{1.40}                   & 2.40                             &  & 2.75          & 2.20          & \textbf{1.05} &        &        &        &        &  \\
		wdbc                              & 2.65                            & \textbf{1.15}                   & 2.20                             &  & 2.55          & \textbf{1.45} & 2.00          &        &        &        &        &  \\ \cmidrule{1-9}
		\textit{avg. rank}                & 1.91                            & \textbf{1.87}                   & 2.21                             &  & 2.53          & 1.83          & \textbf{1.64} &        &        &        &        & \\ \cmidrule{1-9}
	\end{tabular}
	\caption{Friedman ranking on \textbf{classification datasets} using a \textbf{LogReg} for -P.}
	\label{tab:classification-LR}
\end{table}

\begin{table}[h!]
	\small
	\centering
	\begin{tabular}{@{}lrrrlrrrlrrrl@{}}
		\toprule
		\textbf{Dataset} & \multicolumn{4}{c}{\textbf{LR}}             & \multicolumn{4}{c}{\textbf{KMM}}                                & \multicolumn{4}{c}{\textbf{EKMM}}                                                                       \\ \cmidrule(l){2-13}
		& \textbf{-C} & \textbf{-P} & \textbf{-CP} &  & \textbf{-C} & \textbf{-P} & \textbf{-CP} &                      & \multicolumn{1}{l}{\textbf{-C}} & \multicolumn{1}{l}{\textbf{-P}} & \multicolumn{1}{l}{\textbf{-CP}}  &  \\ \midrule
		abalone                           & 2.67                            & 1.81                            & \textbf{1.52}                    &  & 1.68          & 3.00          & \textbf{1.32} &                      & 2.00                 & 2.06                 & \textbf{1.94}        &  \\
		auto-mpg                          & 2.35                            & 1.84                            & \textbf{1.81}                    &  & \textbf{1.37} & 2.82          & 1.81          &                      & 1.97                 & 2.09                 & \textbf{1.94}        &  \\
		autos                             & 2.58                            & \textbf{1.36}                   & 2.06                             &  & 2.01          & 2.15          & \textbf{1.84} &                      & 2.06                 & \textbf{1.96}        & 1.98                 &  \\
		compr-hw                 & \textbf{1.63}                   & 2.28                            & 2.09                             &  & \textbf{1.67} & 2.47          & 1.86          &                      & 2.03                 & 2.05                 & \textbf{1.92}        &  \\
		resi-v9                    & 2.45                            & \textbf{1.21}                   & 2.34                             &  & \textbf{1.88} & 2.20          & 1.92          &                      & 2.00                 & \textbf{1.95}        & 2.05                 &  \\
		resi-v10                   & 1.68                            & \textbf{1.58}                   & 2.74                             &  & \textbf{1.89} & 2.14          & 1.97          &                      & \textbf{1.93}        & 2.03                 & 2.04                 &  \\
		student-mat                       & 2.23                            & \textbf{1.57}                   & 2.20                             &  & \textbf{1.48} & 2.94          & 1.58          &                      & \textbf{1.86}        & 2.18                 & 1.96                 &  \\
		student-por                       & 2.79                            & \textbf{1.42}                   & 1.79                             &  & \textbf{1.60} & 2.80          & \textbf{1.60} &                      & 2.03                 & 2.08                 & \textbf{1.89}        &  \\
		ticdata                           & 2.66                            & 1.77                            & \textbf{1.57}                    &  & 2.14          & 2.66          & \textbf{1.20} &                      & 2.04                 & 2.28                 & 1.68                 &  \\
		wine-q-red                  & 2.81                            & 2.07                            & \textbf{1.12}                    &  & 2.37          & \textbf{1.52} & 2.11          &                      & 2.00                 & \textbf{1.93}        & 2.07                 &  \\
		wine-q-wht                & 2.70                            & \textbf{1.42}                   & 1.88                             &  & \textbf{1.54} & 2.54          & 1.93          &                      & 1.88                 & 2.31                 & \textbf{1.81}        &  \\ \midrule
		\textit{avg. rank}                & 2.41                            & \textbf{1.67}                   & 1.92                             &  & 1.78          & 2.48          & \textbf{1.74} &                      & 1.98                 & 2.08                 & \textbf{1.93}        &  \\ \midrule \midrule
		\textbf{Dataset} & \multicolumn{4}{c}{\textbf{KDE}}             & \multicolumn{4}{c}{\textbf{KLIEP}}                                & \multicolumn{4}{c}{}                                                                       \\ \cmidrule(l){2-9}
		& \textbf{-C} & \textbf{-P} & \textbf{-CP} &  & \textbf{-C} & \textbf{-P} & \textbf{-CP} &                      & \multicolumn{1}{l}{} & \multicolumn{1}{l}{} & \multicolumn{1}{l}{}  &  \\ \cmidrule{1-9}
		abalone                           & 2.17                            & 2.46                            & \textbf{1.37}                    &  & 2.45        & 2.21          & \textbf{1.34} &                      &                      &                      &                      &  \\
		auto-mpg                          & \textbf{1.44}                   & 2.35                            & 2.21                             &  & 2.13        & \textbf{1.80} & 2.07          &        &        &        &        &  \\
		autos                             & \textbf{1.33}                   & 2.64                            & 2.03                             &  & 2.21        & 1.94          & \textbf{1.85} &        &        &        &        &  \\
		comp-hw                 & \textbf{1.34}                   & 2.80                            & 1.86                             &  & 2.76        & \textbf{1.38} & 1.86          &        &        &        &        &  \\
		resi-v9                    & \textbf{1.60}                   & 2.61                            & 1.79                             &  & 2.60        & \textbf{1.66} & 1.74          &        &        &        &        &  \\
		resi-v10                   & \textbf{1.56}                   & 2.44                            & 2.00                             &  & 2.41        & \textbf{1.70} & 1.89          &        &        &        &        &  \\
		student-mat                       & 2.58                            & 2.01                            & \textbf{1.41}                    &  & 2.66        & 1.98          & \textbf{1.36} &        &        &        &        &  \\
		student-por                       & 2.00                            & 2.39                            & \textbf{1.61}                    &  & 2.78        & 1.67          & \textbf{1.55} &        &        &        &        &  \\
		ticdata                           & 2.30                            & \textbf{1.85}                   & \textbf{1.85}                    &  & 2.25        & \textbf{1.80} & 1.95          &        &        &        &        &  \\
		wine-q-red                  & 2.10                            & \textbf{1.59}                   & 2.31                             &  & 2.80        & 1.96          & \textbf{1.24} &        &        &        &        &  \\
		wine-q-wht                & 1.79                            & 2.61                            & \textbf{1.60}                    &  & 2.66        & 2.00          & \textbf{1.34} &        &        &        &        &  \\\cmidrule{1-9}
		\textit{avg. rank}                & 1.84                            & 2.34                            & \textbf{1.82}                    &  & 2.52        & 1.83          & \textbf{1.65} &        &        &        &        &  \\ \cmidrule{1-9}
	\end{tabular}
	\caption{Friedman ranking on \textbf{regression datasets} using a \textbf{SVR} model for -P.}
	\label{tab:regression-SVM}
\end{table}

\begin{table}[h!]
	\small
	\centering
	\begin{tabular}{@{}lrrrlrrrlrrrl@{}}
		\toprule
		\textbf{Dataset} & \multicolumn{4}{c}{\textbf{LR}}             & \multicolumn{4}{c}{\textbf{KMM}}                                & \multicolumn{4}{c}{\textbf{EKMM}}                                                                       \\ \cmidrule(l){2-13}
		& \textbf{-C} & \textbf{-P} & \textbf{-CP} &  & \textbf{-C} & \textbf{-P} & \textbf{-CP} &                      & \multicolumn{1}{l}{\textbf{-C}} & \multicolumn{1}{l}{\textbf{-P}} & \multicolumn{1}{l}{\textbf{-CP}}  &  \\ \midrule
		abalone                           & \textbf{2.00}                   & \textbf{2.00}                   & \textbf{2.00}                    &  & \textbf{2.00} & \textbf{2.00} & \textbf{2.00} &                      & \textbf{2.00}        & \textbf{2.00}        & \textbf{2.00}        &  \\
		auto-mpg                          & \textbf{1.95}                   & 2.10                            & \textbf{1.95}                    &  & 1.80          & 2.55          & \textbf{1.65} &                      & \textbf{1.83}        & 2.23                 & 1.95                 &  \\
		autos                             & \textbf{1.75}                   & 2.13                            & 2.13                             &  & 2.00          & 2.10          & \textbf{1.90} &                      & 2.10                 & \textbf{1.93}        & 1.98                 &  \\
		comp-hw                           & \textbf{1.88}                   & 2.15                            & 1.98                             &  & 2.35          & 2.40          & \textbf{1.25} &                      & \textbf{1.93}        & 2.15                 & \textbf{1.93}        &  \\
		resi-v9                           & 2.05                            & \textbf{1.90}                   & 2.05                             &  & 1.90          & 2.55          & \textbf{1.55} &                      & 2.45                 & \textbf{1.75}        & 1.80                 &  \\
		resi-v10                          & \textbf{1.95}                   & 1.98                            & 2.08                             &  & \textbf{1.35} & 2.70          & 1.95          &                      & \textbf{1.73}        & 2.05                 & 2.23                 &  \\
		student-mat                       & 2.15                            & \textbf{1.93}                   & \textbf{1.93}                    &  & 2.15          & \textbf{1.90} & 1.95          &                      & 2.38                 & 2.13                 & \textbf{1.50}        &  \\
		student-por                       & 2.00                            & \textbf{1.85}                   & 2.15                             &  & \textbf{1.35} & 2.40          & 2.25          &                      & 2.00                 & \textbf{1.70}        & 2.30                 &  \\
		ticdata                           & \textbf{2.00}                   & \textbf{2.00}                   & \textbf{2.00}                    &  & \textbf{2.00} & \textbf{2.00} & \textbf{2.00} &                      & \textbf{2.00}        & \textbf{2.00}        & \textbf{2.00}        &  \\
		wine-q-red                        & 1.90                            & 2.23                            & \textbf{1.88}                    &  & 1.90          & 2.45          & \textbf{1.65} &                      & 2.13                 & 2.50                 & \textbf{1.38}        &  \\
		wine-q-wht                        & \textbf{2.00}                   & \textbf{2.00}                   & \textbf{2.00}                    &  & \textbf{2.00} & \textbf{2.00} & \textbf{2.00} &                      & \textbf{2.00}        & \textbf{2.00}        & \textbf{2.00}        &  \\ \midrule
		\textit{avg.rank}               & \textbf{1.97}                   & 2.02                            & 2.01                             &  & 1.89          & 2.28          & \textbf{1.83} &                      & 2.05                 & 2.04                 & \textbf{1.91}        &  \\ \midrule \midrule
		\textbf{Dataset} & \multicolumn{4}{c}{\textbf{KDE}}             & \multicolumn{4}{c}{\textbf{KLIEP}}                                & \multicolumn{4}{c}{}                                                                       \\ \cmidrule(l){2-9}
		& \textbf{-C} & \textbf{-P} & \textbf{-CP} &  & \textbf{-C} & \textbf{-P} & \textbf{-CP} &                      & \multicolumn{1}{l}{} & \multicolumn{1}{l}{} & \multicolumn{1}{l}{}  &  \\ \cmidrule{1-9}
		abalone                           & 2.03                            & \textbf{2.03}                   & 1.95                             &  & \textbf{2.00} & \textbf{2.00} & \textbf{2.00} &                      &                      &                      &                      &  \\
		auto-mpg                          & 2.50                            & \textbf{1.35}                   & 2.15                             &  & 1.98          & \textbf{1.93} & 2.10          &                      &                      &                      &                      &  \\
		autos                             & 2.40                            & 1.75                            & 1.85                             &  & \textbf{1.95} & 2.05          & 2.00          &                      &                      &                      &                      &  \\
		comp-hw                           & 2.30                            & 2.15                            & \textbf{1.55}                    &  & \textbf{1.88} & 2.23          & 1.90          &                      &                      &                      &                      &  \\
		resi-v9                           & 2.70                            & \textbf{1.60}                   & 1.70                             &  & \textbf{1.98} & 2.05          & 1.98          &                      &                      &                      &                      &  \\
		resi-v10                & 2.85                            & \textbf{1.25}                   & 1.90                             &  & 2.10          & 2.18          & \textbf{1.73} &                      &                      &                      &                      &  \\
		student-mat                       & 2.95                            & \textbf{1.30}                   & 1.75                             &  & 2.05          & \textbf{1.98} & \textbf{1.98} &                      &                      &                      &                      &  \\
		student-por                       & 2.90                            & \textbf{1.40}                   & 1.70                             &  & 2.25          & 1.93          & \textbf{1.83} &          &          &          &          &  \\
		ticdata                           & 2.03                            & 2.03                            & \textbf{1.94}                    &  & \textbf{2.00} & \textbf{2.00} & \textbf{2.00} &          &          &          &          &  \\
		wine-q-red                        & 1.60                            & \textbf{1.45}                   & 2.95                             &  & 2.10          & 2.08          & \textbf{1.83} &          &          &          &          &  \\
		wine-q-wht                        & \textbf{2.00}                   & \textbf{2.00}                   & \textbf{2.00}                    &  & \textbf{2.00} & \textbf{2.00} & \textbf{2.00} &          &          &          &          &  \\\cmidrule{1-9}
		\textit{avg. rank}                & 2.39                            & \textbf{1.66}                   & 1.95                             &  & 2.03          & 2.04          & \textbf{1.94} &          &          &          &          & \\ \cmidrule{1-9}
	\end{tabular}
	\caption{Friedman ranking on \textbf{regression datasets} using \textbf{Ridge} for -P.}
	\label{tab:regression-LR}
\end{table}

\begin{table}[h!]
{\small
	\centering
	\begin{tabular}{@{}lllllll@{}}
		\toprule
		&& \textbf{LR} & \textbf{KMM} & \textbf{EKMM} & \textbf{KDE} & \textbf{KLIEP} \\ \midrule
		\multirow{2}{*}{class.} & SVM       & P           & C            & CP            & CP           & CP             \\
		& LogReg & C           & CP           & CP            & P            & CP             \\ \midrule
		\multirow{2}{*}{regr.} & SVR           & P           & CP           & CP            & CP           & CP             \\
		 & Ridge         & C           & CP           & CP            & P            & CP             \\ \bottomrule
	\end{tabular}
	\caption{Summary of the best approach for each of the different systems on artificial datasets.}
	\label{tab:resumen-artificial}}
\end{table}

Table \ref{tab:resumen-artificial} summarizes the performance of the approaches for classification and regression. In general, the systems seem to benefit from taking both covariate and prediction information for computing the importance, or just predictions in case of KDE. Unfortunately, there exist some exceptions. One case is LR, for which only taking just predictions gives the best results for SVM and SVR and taking just covariates for LogReg and Ridge. Another exception occurs for KMM in classification using SVM, for which the best performance is reached when just the covariates are taken. However, this anomaly is just due to the transfusion dataset and the high difference between -C and -CP. EKMM and KLIEP offer the best performance when both kind of information is considered whatever the system used and either in classification or regression. The confidence degree obtained at the significant level $\alpha = 0.05$ using the Nemenyi test \citep{hollander2013nonparametric} is $1.0483$ on classification and $0.9995$ on regression. This means that there exists a significant different at level $\alpha = 0.05$ when doing pairwise comparison in the rankings if their absolute difference is bigger than the confidence degree. Taking into account this fact, the results do not show significative differences.

KMM-P considerably worsens with regard to KMM-C. Since KMM directly computes the importance and does not estimate any type of density or probability distribution with the covariates, it is reasonable that using target information instead of the covariates yields worse results, since the way this method uses covariate information does not give room to consider the covariate and target information equivalent. This assumption is further proved by means of using the target information alongside the covariates, that is, KMM-CP, which generates an improvement with regard to KMM-C. Hence, this confirms the usefulness of the predictions, even feed to systems when covariates are also crucial in the computation of the importance. This behaviour remains for the ensemble version EKMM.

Concerning KDE, the error estimation seems to worse when only target information is used, that is, KDE-P. Nonetheless, it also seems to improve massively for some datasets in both classification and regression problems. This could be a result of the quality of the predictions obtained by the model, a hypothesis that is tested later on by changing SVM into LogReg. When it comes to KDE-CP, the results are better than those of KDE-C, which indicates that KDE benefits from target information, as bad as the predictions used may be.

Nonetheless, as expected, the best performance is reached with the more promising procedure \citep{sugiyama2012machine}, that is, KLIEP. This system is able to get information from both covariates and predictions, being the best option taking both.

These conclusions confirm our hypothesis that target information would improve the importance estimation, both on its own and when used in combination with the covariates information.

\begin{table*}[h!]
  \small
	\centering
	\begin{tabular}{@{}lrrrlrrrlrrrl@{}}
	\toprule
\multirow{2}{*}{\textbf{Dataset}} & \multicolumn{4}{c}{\textbf{LR}}             & \multicolumn{4}{c}{\textbf{KDE}}                                & \multicolumn{4}{c}{\textbf{EKMM}}                                                                       \\ \cmidrule(l){2-13}
	                                  & \textbf{-C} & \textbf{-P} & \textbf{-CP} &  & \textbf{-C} & \textbf{-P} & \textbf{-CP} &                      & \multicolumn{1}{l}{\textbf{-C}} & \multicolumn{1}{l}{\textbf{-P}} & \multicolumn{1}{l}{\textbf{-CP}}  &  \\ \midrule
p2006a-2006b        & 3.00                            & \textbf{1.00}                   & 2.00                             &  & 3.00          & 1.90          & \textbf{1.10} &  & 2.00          & 2.10          & \textbf{1.90} &  \\
p2006b-2007a        & 3.00                            & \textbf{1.40}                   & 1.60                             &  & 3.00          & 2.00          & \textbf{1.00} &  & \textbf{1.70} & 2.50          & 1.80          &  \\
p2007a-2007b        & 3.00                            & 2.00                            & \textbf{1.00}                    &  & 3.00          & \textbf{1.00} & 2.00          &  & \textbf{1.70} & 2.50          & 1.80          &  \\
p2007b-2008a        & 3.00                            & \textbf{1.00}                   & 2.00                             &  & 3.00          & \textbf{1.00} & 2.00          &  & \textbf{1.80} & \textbf{1.80} & 2.40          &  \\
p2008a-2008b        & 3.00                            & 2.00                            & \textbf{1.00}                    &  & 3.00          & \textbf{1.00} & 2.00          &  & 2.30          & 2.20          & \textbf{1.50} &  \\
p2008b-2009a        & 2.60                            & \textbf{1.00}                   & 2.40                             &  & 3.00          & 2.00          & \textbf{1.00} &  & 2.20          & 2.30          & \textbf{1.50} &  \\
p2009a-2009b        & 3.00                            & 2.00                            & \textbf{1.00}                    &  & 3.00          & \textbf{1.00} & 2.00          &  & 2.10          & 2.10          & \textbf{1.80} &  \\
p2009b-2010a        & \textbf{1.00}                   & 3.00                            & 2.00                             & & 3.00          & \textbf{1.00} & 2.00          &  & \textbf{2.00} & \textbf{2.00} & \textbf{2.00} &  \\
p2010a-2010b        & 3.00                            & 2.00                            & \textbf{1.00}                    &  & 3.00 				  & \textbf{1.00} & 2.00          &  & \textbf{1.60} & 2.40          & 2.00          &  \\
p2010b-2011a        & 2.00                            & 3.00                            & \textbf{1.00}                    &  & 3.00          & 2.00          & \textbf{1.00} &  & 2.00          & \textbf{1.90} & 2.10          &  \\
p2011a-2011b        & 3.00                            & 2.00                            & \textbf{1.00}                    &  & 3.00          & \textbf{1.00} & 2.00          &  & 2.10          & \textbf{1.40} & 2.50          &  \\
p2011b-2012a        & 3.00                            & 2.00                            & \textbf{1.00}                    &  & 3.00          & 2.00          & \textbf{1.00} &  & 1.90          & 2.50          & 1.60          &  \\
p2012a-2012b        & 3.00                            & 2.00                            & \textbf{1.00}                    &  & 3.00          & \textbf{1.00} & 2.00          &  & \textbf{1.80} & 1.90          & 2.30          &  \\
p2012b-2013a        & 3.00                            & \textbf{1.00}                   & 2.00                             &  & 3.00          & 1.80          & \textbf{1.20} &  & 2.00          & 2.10          & \textbf{1.90} &  \\
p2013a-2013b        & 3.00                            & 2.00                            & \textbf{1.00}                    &  & 3.00          & \textbf{1.00} & 2.00          &  & 2.10          & 2.10          & \textbf{1.80} &  \\ \midrule
\textit{avg. ranks} & 2.77                            & 1.83                            & \textbf{1.40}                    & $\S$ & 3.00          & \textbf{1.38} & 1.62          & $\S$ & 1.95          & 2.12          & \textbf{1.93} & \\\bottomrule
\end{tabular}
	\caption{Friedman ranking on \textbf{plankton datasets} using a \textbf{SVR} model for -P.}
	\label{tab:plankton-SVM}
\end{table*}

\begin{table*}[h!]
	\small
	\centering
	\begin{tabular}{@{}lrrrlrrrlrrrl@{}}
	\toprule
\multirow{2}{*}{\textbf{Dataset}}  & \multicolumn{4}{c}{\textbf{LR}}             & \multicolumn{4}{c}{\textbf{KDE}}                                & \multicolumn{4}{c}{\textbf{EKMM}}                                                                       \\ \cmidrule(l){2-13}
        & \textbf{-C} & \textbf{-P} & \textbf{-CP} &  & \textbf{-C} & \textbf{-P} & \textbf{-CP} &                      & \multicolumn{1}{l}{\textbf{-C}} & \multicolumn{1}{l}{\textbf{-P}} & \multicolumn{1}{l}{\textbf{-CP}}  &  \\ \midrule
				p2006a-2006b       & 3.00          & 2.00          & \textbf{1.00} &  & 3.00 & 2.00 & \textbf{1.00} &  & 2.00          & \textbf{1.00} & 3.00          &  \\
				p2006b-2007a       & \textbf{1.00} & 2.00          & 3.00          &  & 3.00 & 2.00 & \textbf{1.00} &  & \textbf{1.00} & 2.00          & 3.00          &  \\
				p2007a-2007b       & 3.00          & 2.00          & \textbf{1.00} &  & 3.00 & 2.00 & \textbf{1.00} &  & 3.00          & 2.00          & \textbf{1.00} &  \\
				p2007b-2008a       & 3.00          & 2.00          & \textbf{1.00} &  & 3.00 & 2.00 & \textbf{1.00} &  & \textbf{1.00} & 3.00          & 2.00          &  \\
				p2008a-2008b       & 3.00          & 2.00          & \textbf{1.00} &  & 3.00 & 2.00 & \textbf{1.00} &  & 2.00          & 3.00          & \textbf{1.00} &  \\
				p2008b-2009a       & 2.00          & \textbf{1.00} & 3.00          &  & 3.00 & 2.00 & \textbf{1.00} &  & \textbf{1.00} & 3.00          & 2.00          &  \\
				p2009a-2009b       & 3.00          & 2.00          & \textbf{1.00} &  & 3.00 & 2.00 & \textbf{1.00} &  & \textbf{1.00} & 2.00          & 3.00          &  \\
				p2009b-2010a       & \textbf{1.00} & 2.00          & 3.00          &  & 3.00 & 2.00 & \textbf{1.00} & & 3.00          & 2.00          & \textbf{1.00} &  \\
				p2010a-2010b       & 3.00          & 2.00          & \textbf{1.00} &  & 3.00 & 2.00 & \textbf{1.00} &  & 3.00          & \textbf{1.00} & 2.00          &  \\
				p2010b-2011a       & 2.00          & 3.00          & \textbf{1.00} &  & 3.00 & 2.00 & \textbf{1.00} &  & 3.00          & \textbf{1.00} & 2.00          &  \\
				p2011a-2011b       & \textbf{1.00} & 3.00          & 2.00          &  & 3.00 & 2.00 & \textbf{1.00} &  & 2.00          & \textbf{1.00} & 3.00          &  \\
				p2011b-2012a       & 3.00          & 2.00          & \textbf{1.00} &  & 3.00 & 2.00 & \textbf{1.00} &  & 3.00          & \textbf{1.00} & 2.00          &  \\
				p2012a-2012b       & 3.00          & 2.00          & \textbf{1.00} &  & 3.00 & 2.00 & \textbf{1.00} &  & \textbf{1.00} & 2.00          & 3.00          &  \\
				p2012b-2013a       & 3.00          & 2.00          & \textbf{1.00} &  & 3.00 & 2.00 & \textbf{1.00} &  & \textbf{1.00} & 3.00          & 2.00          &  \\
				p2013a-2013b       & 3.00          & 2.00          & \textbf{1.00} &  & 3.00 & 2.00 & \textbf{1.00} &  & 3.00          & 2.00          & \textbf{1.00} &  \\ \midrule

\textit{avg.ranks} & 2.47       & 2.07       & \textbf{1.47} & $\S$  & 3.00 & 2.00 & \textbf{1.00} & $\S$ & 2.00       & \textbf{1.93} & 2.07       & \\ \bottomrule
\end{tabular}
\caption{Friedman ranking on \textbf{plankton datasets} using \textbf{Ridge} for -P.}
	\label{tab:plankton-LR}
\end{table*}

\begin{table}[h!]
	\centering
	\begin{tabular}{@{}llll@{}}
		\toprule
		& \textbf{LR} & \textbf{KDE} & \textbf{EKMM} \\ \midrule
		SVR   & CP          & P            & CP            \\
		Ridge & CP          & CP           & P             \\ \bottomrule
			\end{tabular}
		\caption{Summary of the best approach for each of the different systems on the plankton datasets.}
		\label{tab:resumen-plankton}
\end{table}

Let us comment the results for the plankton dataset. Tables \ref{tab:plankton-SVM} and \ref{tab:plankton-LR} show the results for plankton datasets. Since it is a regression task, SVR and Ridge are applied. In this case, and up until now only experiments using LR, KDE and EKMM were possible to be carried out. The existing implementation for KMM has reported memory errors in the kernel computation, whereas the existing implementation for KLIEP takes prohibitive time in optimizing the bandwidth parameter. The confidence degree using Nemenyi test is $0.8559$ for a significant level $\alpha= 0.05$, so the rankings with a higher step to this degree between them will be considered significantly different. The symbol $\S$ in the tables indicates if there is significant differences of the best method. This is the case of $LR$ and $KDE$ In this case, LR benefits from target information, since both LR-P and LR-CP improve the performance with regard to LR-C. KDE also improves with target information, although it seems best if this is used without the covariates, as KDE-P is better than KDE-CP. Finally, and as seen before, EKMM seems to worsen when using only predictions, (EKMM-P is worse than EKMM-C), but adding target information to the covariates works better, as EKMM-CP is the best out of the three EKMM approximations. Also, Table \ref{tab:resumen-plankton} summarizes the performance of the approaches. Looking at this table, the main conclusion is that taking predictions improves the performance of the systems, especially if both covariates and predictions are taken into account.

\section{Conclusions}
Covariate shift, although really prominent in the real-world, has not been studied in the literature as other situations free of dataset shift. Accurate error estimation is one of the key challenges in machine learning that have been widely studied. However, traditional methods for this purpose do not longer work in a covariate shift context.

Although several methods have already been proposed in this field, none of them takes into account the target information, merely focusing on covariate information. The research reported in this paper arises on the basis of including target information will lead to a potential improvement in the error estimation process. In this direction, this paper takes some existing methods  for the error estimation process in covariate shift that are based on the concept of importance computation and checks if an improvement takes place when they are fed target information, both on its own in lieu of covariates, and alongside them. This implies redefining the importance in a more general way, mapping the covariates into a feature space that takes the target information. The basic Logistic Regression (LR), the Kernel Mean Matching (KMM), the Ensemble Kernel Mean Matching (EKMM), the Kernel Density Estimation (KDE), and the Kullback-Leibler Importance Estimation Procedure (KLIEP) were taken for comparison. The target information has been included through model predictions, since one can assume the existence of a correlation between covariates and targets even in situations under covariate shift. The methods are tested over both artificial classification and regression datasets in addition to a plankton classification dataset, which was the origin of this research.

The results reveal, in one way, what it was already expected. In general, the best results are reached when both covariates and target information in form of predictions are taken into account. In case of plankton dataset the scores are statistically significant. Some systems that directly computes the density function, as LR or KDE, seems to perform better when just target information is taken, what makes think that they do not deal properly with possible noise or redundancy in the covariates.

The experiments over the plankton dataset hint at a solution for error estimation when high-dimensionality becomes a drawback. Introducing and exploiting the information contained in the targets instead of just that of the covariates allows for i) a more memory efficient estimation of the importance and ii) improving accuracy of the error estimation. Moreover, and given there is enough resources in terms of memory and time, using target information along the covariates further improves the accuracy of the error estimation.

As future work, and concerning the redefinition of importance mapping the covariates into a feature space, more proposals different from the predictions can be explored in a attempt to get an improvement in the importance computation. Also, a straightforward and immediate task would be to obtain more accurate predictions using, for instance, specific models for each dataset or even simply tuning clue parameters. The difference in performance of some methods when changing the predictions model show that the quality of them could help to improve even more the accuracy of the error estimation. Also, creating more efficient implementations in order to obtain more scalable methods for high dimensionality applications, such as plankton, could also be another future research line to follow.

\section*{Acknowledgments}\label{sec:Acknowledgments}
This research has been partially supported by the Spanish Ministerio de Ciencia e Innovación through the grant PID2019-110742RB-I00.





\bibliographystyle{apalike}
\biboptions{authoryear,semicolon}




\bibliography{mybibliography}

\end{document}